\documentclass[final]{OAGM}

\OAGMarXiv{1505.01065}

\usepackage[pdftex]{graphicx}
\usepackage{graphicx,fancyhdr}
\usepackage{amssymb}
\usepackage{amsmath}
\usepackage{color}
\usepackage{url}
\usepackage{hyperref}
\usepackage{algorithm}

\title{Object Modelling with a Handheld RGB-D~Camera}

\author{Aitor Aldoma, Johann Prankl, Alexander Svejda and Markus Vincze\\
         Automation and Control Institute, Vienna University of Technology, Austria\\
        {\tt\footnotesize \{prankl, aldoma, vincze\}@acin.tuwien.ac.at} }

\begin{document}
\maketitle

\begin{abstract}
  This work presents a flexible system to reconstruct 3D models of objects captured with an \mbox{RGB-D} sensor. A major advantage of the method is that our reconstruction pipeline allows the user to acquire a full 3D model of the object. This is achieved by acquiring several partial 3D models in different sessions that are automatically merged together to reconstruct a full model. In addition, the 3D models acquired by our system can be directly used by state-of-the-art object instance recognition and object tracking modules, providing object-perception capabilities for different applications, such as human-object interaction analysis or robot grasping. The system does not impose constraints in the appearance of objects (textured, untextured) nor in the modelling setup (moving camera with static object or a turn-table setup). The proposed reconstruction system has been used to model a large number of objects resulting in metrically accurate and visually appealing 3D models.
\end{abstract}

\section{Introduction}

The availability of commodity \mbox{RGB-D} sensors, combined with several advances in 3D printing technology, has sparked a renewed interest in software tools that enable users to digitize objects easily and most importantly, at low economical costs. However, being able to accurately reconstruct 3D object models has not only applications among modelling or 3D printing aficionados, but also in the field of robotics. For instance, the knowledge in form of 3D models can be used for object instance recognition, enabling applications such as autonomous grasping or object search under clutter and occlusions.

While numerous reconstruction tools exist to capture 3D models of environments, only a few of them focus on the reconstruction of individual objects. This can be partially ascribed to the difference in scale between objects (e.g. household objects) and larger environments (e.g. rooms or buildings, usually the focus of SLAM systems), the need to subtract the object of interest from the rest of the environment as well as other nuisances that make object reconstruction a challenging problem. For example, the requirement of full 3D models is ignored by most reconstruction systems.
\begin{figure}[t]
  \begin{center}
    \includegraphics[width=.55\textwidth]{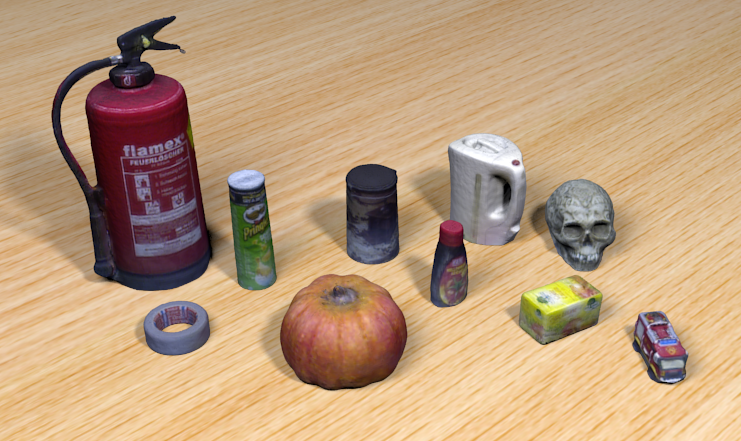}
    \caption{Virtual scene recreated with some of the 3D models reconstructed by the proposed modelling tool.} \vspace{-0.7cm}
    \label{fig:awesomeness}
  \end{center}
\end{figure}

Aiming at addressing the aforementioned challenges into an integrated system as well as at enabling recognition and tracking, we propose in this work an integrated object reconstruction pipeline that (i) is easy to use, (ii) results in metrically accurate and visually appealing models, (iii) does not make assumptions of the kind of objects being modelled\footnote{As long as they can be sensed by RGB-D sensors} (iv) reconstructs full 3D models by merging partial models acquired in different sessions and (v) is able to export object models that can be seamlessly integrated into object recognition and tracking modules without any additional hassle. We will release our modelling and object perception systems to enable this.

In the remainder of this paper, we present the different modules of the system, focusing on those with novel characteristics or that are crucial to the robustness of the overall system. Because the evaluation of complex pipelines like the one proposed in this paper is always a major challenge, we compare the fidelity of the end result (i.e. 3D models) obtained with our system with their counterparts reconstructed using a precise laser scanner. This quantitative comparison shows that the reconstructed models are metrically accurate: the average error ranging between one and two millimetres.

\vspace{-0.2cm}
\section{Related work}
\vspace{-0.2cm}
The proposed framework covers a broad variety of methods including registration, object segmentation, surface reconstruction, texturing and supported applications such as object tracking and object recognition. In this section we focus on related work of the core methods necessary for object modelling: camera tracking, point cloud registration and surface modelling. 

Using interest points is one of the most popular ways of finding correspondences in image pairs enabling the registration of RGB-D frames. For example Endres~et~al.~\cite{endres2014mapping} developed a Visual SLAM approach which is able to track the camera pose and register point clouds in large environments. Loop closing and a graph based optimization method are used to compensate the error accumulated during camera tracking. 
Especially for re-localization we also rely on interest points. In contrast to Endres~et~al.~\cite{endres2014mapping} we develop a LK-style tracking approach which is able to minimize the drift, enabling to create models for tracking and recognition without the necessity of an explicit loop closing.

Another type of methods is based on the well established Iterative Closest Point (ICP) algorithm~\cite{huber2003fully,fantoni2012accurate,krainin2011manipulator}. Huber~et~al.~\cite{huber2003fully} as well as Fantoni~et~al.~\cite{fantoni2012accurate} focus on the registration of unordered sets of range images.
In~\cite{krainin2011manipulator} the authors propose a robotic in-hand object modelling approach where the object and the robotic manipulator are tracked with an articulated ICP variant. 

While the above systems generate sparse representations, namely point clouds, the celebrated approach of Izadi~et~al.~\cite{izadi2011acm} uses a truly dense representation based on signed distance functions. 
Since then, several extensions have appeared. While the original Kinect Fusion~\cite{izadi2011acm} relies on depth data Kehl~et~al.~\cite{kehl2014bmvc} introduce a colour term and is like our proposal able to register multiple modelling sessions. However, \cite{kehl2014bmvc} relies on sampling the rotational part of the pose space in order to provide initial approximations to their registration method. Instead, we use features and stable planes to attain initial alignments effectively reducing computational complexity. A direct approach for registration is proposed in Bylow~et~al.~\cite{bylow2013realtime}. They omit ICP and directly optimize the camera poses using the SDF-volume. Furthermore, the first commercial scanning solutions such as ReconstructMe, itSeez3D and CopyMe3D became available.

In summary, we propose a robust and user-friendly approach which is flexible enough to adapt to different user requirements and is able to generate object models for tracking and recognition. 

\vspace{-0.2cm}
\section{System overview}
\vspace{-0.2cm}
\begin{figure}[t]
  \begin{center}
    \includegraphics[width=.7\textwidth]{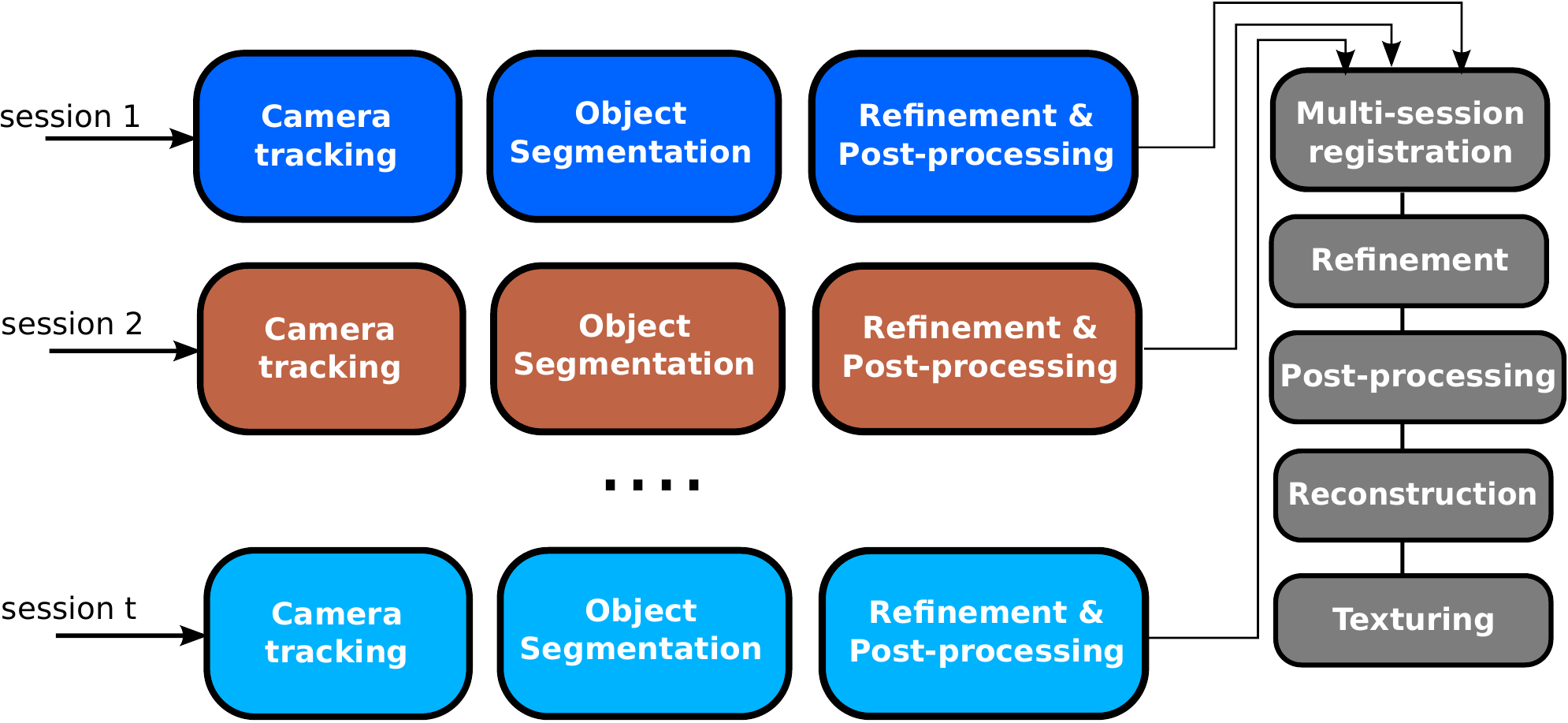}
    \caption{ Pictorial overview of the proposed object modelling pipeline.}
    \vspace{-0.7cm}
    \label{fig:overview}
  \end{center}
\end{figure}

Approaches for object modelling typically involve accurate camera tracking, object segmentation and, depending on the application, a post-processing step which includes pose refinement and eventually surface reconstructing and texturing. Concerning camera tracking, we use a visual odometry based on tracked interest points. If the object itself is texture-less, we rely on background texture (e.g. by adding a textured sheet of paper on the supporting surface) in order to successfully model these kind of objects. The camera positions are refined by means of bundle adjustment as well as an accurate multi-view ICP approach. 
Segmentation of the object of interest from the background is attained by a multi-plane detection and a smooth clustering approach offering object hypotheses to be selected by the user. Alternatively, a simple bounding box around the object can be used to define a region of interest from which the object is easily singled out from the background.
If a complete model (i.e. including the bottom and self-occluded parts) is desired, a registration approach is proposed to automatically align multiple sequences. Finally, our system includes a post-processing stage to reduce artefacts coming from noisy observations as well as a surface reconstruction and texturing module to generate dense and textured meshes. A schematic representation of the modelling pipeline is depicted in Figure \ref{fig:overview}. The individual steps including novel aspects of the system are explained in more detail in the following sections.

\vspace{-0.2cm}
\subsection{Registration and segmentation} \label{sec:tracking}
\vspace{-0.1cm}
A key component for the reconstruction of 3D models is the ability to accurately track the camera pose with respect to the object of interest. 
\begin{figure}[t]
  \begin{center}
    \includegraphics[width=.6\textwidth]{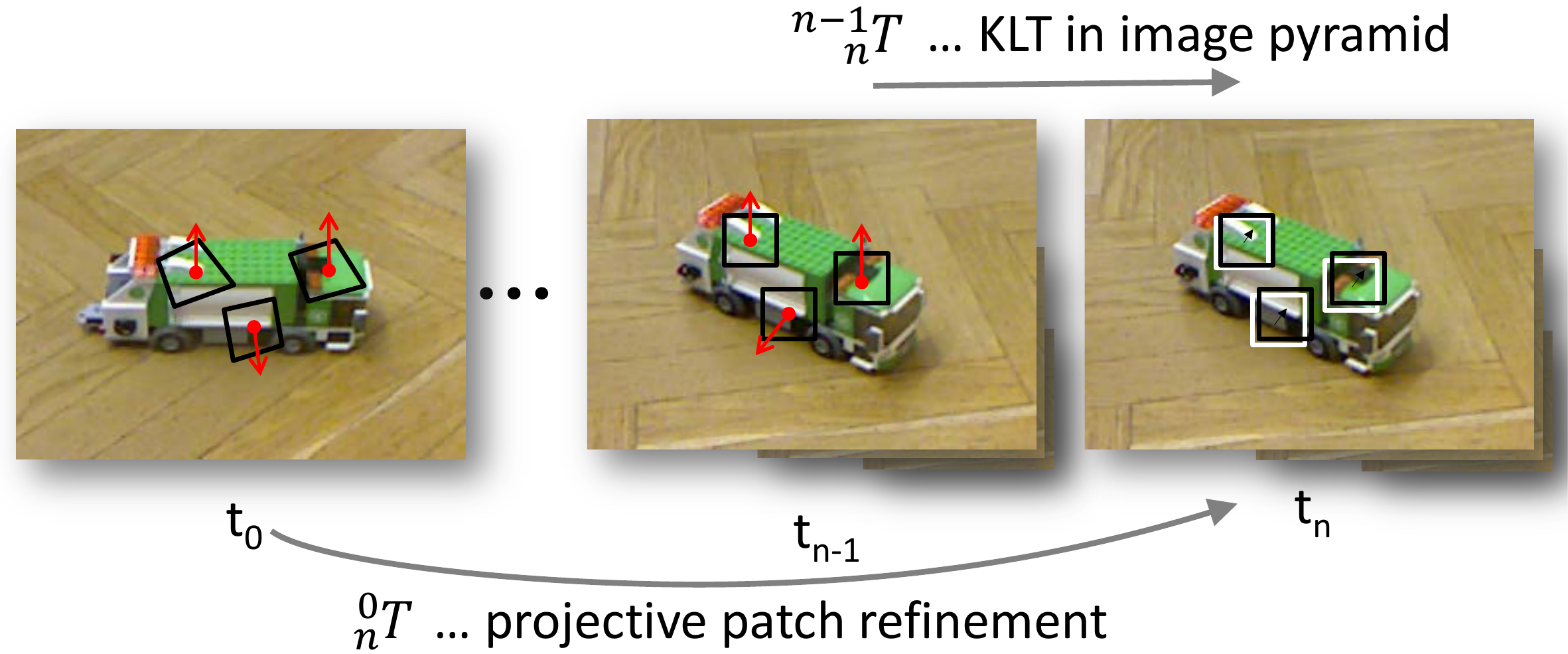}
    \caption{ Camera tracking including a frame by frame visual odometry and a projective patch refinement from keyframe to frame. }
    \vspace{-0.6cm}
    \label{fig:camera_tracking}
  \end{center}
\end{figure}

The proposed approach combines frame by frame tracking based on a KLT-tracker~\cite{tomasi1991detection} and a keyframe based refinement step. Once a keyframe is created, normals are estimated and in combination with the pose hypothesis a locally correct patch warping (homography) from the keyframe to the current frame is performed. The patch location is then refined in an additional KLT-style refinement step. Keyframes are stored in case the camera motion is lager than a threshold. Finally, the camera pose is estimated using the depth information from the organized RGB-D frames. Fig.~\ref{fig:camera_tracking} depicts the tracking approach, where $T$ indicates the pose transformation computed from tracked points. 

This camera tracking framework is already capable of modelling complete scenes in real-time. If one wants to reconstruct individual objects an additional manual interaction is necessary. As already mentioned, we provide two options to segment objects, namely
an interactive segmentation approach which relies on multi-plane detection and smooth object segmentation, and segmentation based on a tracked region of interest (ROI).

The result of this stage is a set of indices $\mathcal{I} = \{ I^1, ..., I^{n_i} \}$, $I^k$ indicating the pixels of $K^k$ containing the object of interest. An initial point cloud of the object can be reconstructed as $\mathcal{P} = \bigcup_{k=1:n} T^k \left ( K^k[I^k] \right)$ where $K[\cdot]$ indicates the extraction of a set of indices from a keyframe.

\vspace{-0.2cm}
\subsection{Multi-view refinement}
\vspace{-0.1cm}
While the visual odometry presented in Section~\ref{sec:tracking} has proven to be sufficiently accurate for the envisioned scenario, the concatenation of several transformations inevitably results in a certain drift in the overall registration. Aiming at mitigating this undesirable effect we provide two alternative mechanisms to reduce the global registration error. 
On one hand, the framework allows to perform bundle-adjustment in order to reduce the re-projection error of correspondences used during camera tracking. On the other hand, the system is equipped with the multi-view Iterative Closest Point introduced in~\cite{fantoni2012accurate} that globally reduces the registration error between overlapping views by iteratively adapting the transformation between camera poses. While multi-view ICP is considerable slower than bundle-adjustment, its application is not constrained to objects with visual features and due to its dense nature, results in more accurate registrations.

\vspace{-0.1cm}
\section{Post-processing}
\begin{figure}[t]
  \begin{center}
    \includegraphics[width=.6\textwidth]{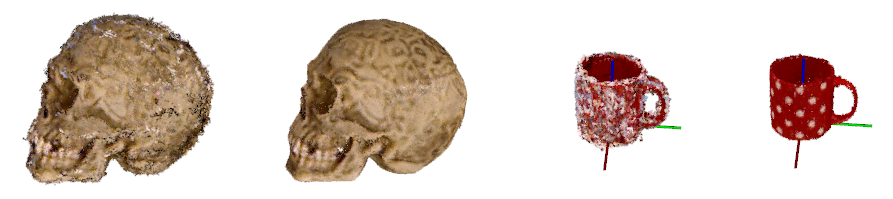}
    \caption{ Effects of the post-processing stage on the reconstruction results.}
    \vspace{-0.5cm}
    \label{fig:post_processing}
  \end{center}
\end{figure}

\vspace{-0.1cm}
The methods presented so far have been designed to be robust to noise and sensor nuisances. However, such artefacts are present in the data and a post-processing stage is required to remove them in order to obtain a visually appealing and accurate model. The techniques within this section provide a pleasant reconstruction by removing these artefacts from the underlying data. Figure \ref{fig:post_processing} visualizes the improvement on the final reconstruction after the post-processing stage. Please note, that the methods herein, do not change the alignment results obtained during the registration process.

\subsection{Noise model}
\vspace{-0.1cm}
In \cite{NguyenIL12}, the authors study the effect of surface-sensor \emph{distance} and \emph{angle} on the data. They obtain axial and lateral noise distributions by varying the aforementioned two variables and show how to include the derived noise model into Kinect Fusion \cite{izadi2011acm} to better accommodate noisy observations in order to reconstruct thin and challenging areas. 

In particular, for object modelling, \emph{surface-sensor angle} is more important than distance, since the later can be controlled and kept at an optimal range (i.e., one meter or closer). Following \cite{NguyenIL12}, we observe that:
\begin{itemize}
  \vspace{-0.1cm}
 \item Data quickly deteriorates when the angle between the sensor and the surface gets above $60$ degrees.
  \vspace{-0.1cm}
 \item Lateral noise increases linearly with distance to the sensor. It results in jagged edges close to depth discontinuities causing the measured point to jump between foreground and background.
 Observe the white points on the left instances of reconstructed models in Figure \ref{fig:post_processing} coming from the plane on the background where the objects are standing.
\end{itemize}
\vspace{-0.1cm}

From these two observations, we propose a simple noise model suited for object modelling that results in a significant improvement on the visual quality of the reconstruction. Let $\mathcal{C}=\{p_i\}$ represent a point cloud in the sensor reference frame, $\mathcal{N}=\{n_i\}$ the associated normal information and $\mathcal{E}=\{e_i\}$, $e_i$ being a boolean variable indicating whether $p_i$ is located at a depth discontinuity or not. $w_i$ is readily computed as follows:

\begin{equation}
\left.
\begin{aligned}
w_i = \left(1 - \frac{\theta - \theta_{max}}{90 - \theta_{max}} \right) \cdot \left(1 - \frac{1}{2} exp^{\frac{d_i^2}{\sigma_L^2}} \right)
\end{aligned}
\right.
\label{eq:noise_model_weight}
\end{equation}

where $\theta$ represents the angle between $n_i$ and the sensor, $\theta_{max}=60^\circ$, $d_i=\vert\vert p_i - p_j \vert\vert_2$ ($p_j$ being the closest point with $e_j=true$) and $\sigma_L=0.002$ represents the lateral noise sigma. Lateral noise is almost constant up to a certain angle, $\frac{\theta - \theta_{max}}{90 - \theta_{max}}=0$ if $\theta < \theta_{max}$. 
Because the selected keyframes present a certain overlap, we improve the final point cloud by averaging good  (based on the noise model weights) observations that lie on the same actual surface as well as by removing inconsistent observations. 

\vspace{-0.1cm}
\section{Multi-session alignment}
\vspace{-0.1cm}
In this section, we discuss the proposed techniques to automatically align multiple sessions into a consistent 3D model. Please note that since the configuration of the object has been changed with respect to its surroundings (e.g. supporting plane), this process needs to rely solely on the information provided by the object. 

Let $\mathcal{P}_{1:t}$ be a set of $t$ partial 3D models obtained by reconstructing the same object in different configurations. 
The goal now is to find a set of transformations that align the different scans, $\mathcal{P}_{1:k}$, into the coordinate system of (without loss of generality) $\mathcal{P}_{1}$. For simplicity, let us discuss first the case where $t=2$. 
In this case, we seek a single transformation aligning $\mathcal{P}_2$ to $\mathcal{P}_1$. To obtain it, we make use of the initial alignments, either provided by a feature based registration~\cite{aldoma2012tutorial} or by exploiting stable planes of objects in combination with ICP~\cite{aldoma2011}.
Each initial alignment is then refined by means of ICP. Because several initial alignments can be provided, we need to define a metric to evaluate the registration quality. The transformation associated with the best registration according to this criteria, will be then the sought transformation. This quality criteria is based on two aspects: (i) amount of points causing free space violation~(FSV) and (ii) amount of overlap. Recall from~\cite{Huber_2001_3886} that the FSV ratio between two point clouds is efficiently computed as the ratio of the number of points of the first cloud in front of the surface of the second cloud over the number of points in the same surface. Intuitively, we would like on one hand to favour transformations causing a small amount of free space violations (indicating consistent alignments) and on the other hand, to favour alignments that present enough overlap to compute an accurate transformation.

If $t\ge3$, we repeat the process above for all pairs $(\mathcal{P}_i$,$\mathcal{P}_j)_{i>j}$. Then, we create a weighted graph with $k$ vertices and edges between vertices  including the best transformation aligning $(\mathcal{P}_i$,$\mathcal{P}_j)$ together with the computed quality measure. Then, a unique registration of all partial models is obtained by computing the MST of the graph and appropriately concatenating the transformations found at the edges of the tree when traversing from $\mathcal{P}_i$ to $\mathcal{P}_1$. After all partial models have been brought into alignment, the multi-view refinement process as well as the post-processing stage previously described may be executed for further accuracy.
\begin{figure}[tpb]
  \begin{center}
    \includegraphics[width=0.2\textwidth]{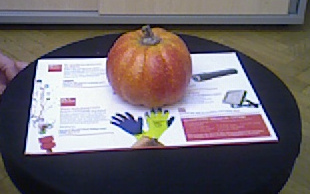}
    \includegraphics[width=0.2\textwidth]{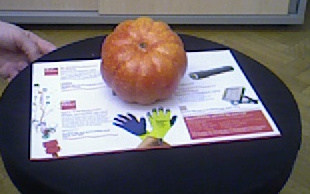}
    \includegraphics[width=0.34\textwidth]{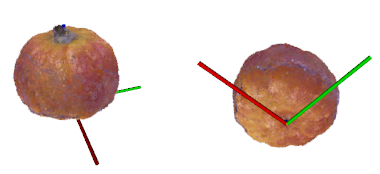} \\
    \includegraphics[width=0.2\textwidth]{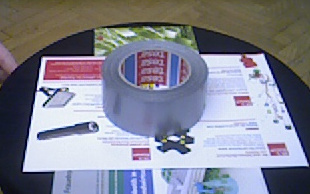}
    \includegraphics[width=0.2\textwidth]{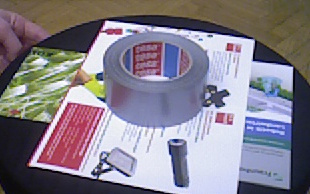}
    \includegraphics[width=0.34\textwidth]{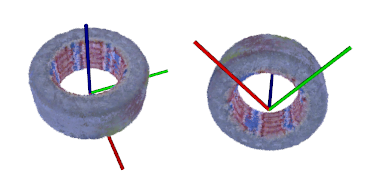}
    \caption{ Examples of successful alignments between sessions by means of stable planes. The objects do not present enough unique features matchable across sessions to enable registration using features.}
    \label{fig:pumpkin_tape}
  \end{center}
\end{figure}
Figure \ref{fig:pumpkin_tape} shows two examples where the objects do not have enough features to be matched across sessions (due to its repetitive structure) that are however correctly aligned using stable planes and ICP.

\section{Experimental results}
\label{sec:eval}

\begin{figure}[tpb]
  \begin{center}
    \includegraphics[height=30mm]{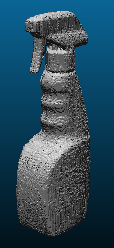}
    \includegraphics[height=30mm]{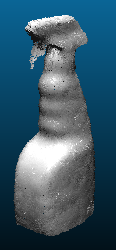}
    \includegraphics[height=30mm]{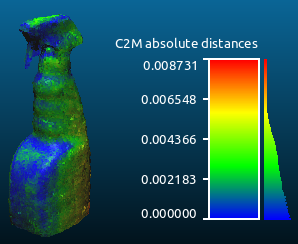}
    \caption{ Distance from reconstructed point clouds against laser scanner model. Distance ($\mu \pm \sigma$): 2.16 $\pm$ 1.53mm}
    \label{fig:comp1}
  \end{center}
\end{figure}

In addition to the qualitative results shown throughout this work, this section evaluates the accuracy of the reconstructed models with respect to models of the same objects acquired with a precise Laser Scanner~\cite{Kasper29052012}.
We assess the quality of our reconstructions by comparing the reconstructed 3D models with their counterparts from the KIT Object Models Web Database~\cite{Kasper29052012}. To do so, we use the CloudCompare software\footnote{\url{http://www.danielgm.net/cc/}} in order to interactively register both instances of the objects and to compute quality metrics. In particular, the error is assessed by computing statistics regarding the closest distance from the reconstructed point cloud to the mesh provided by~\cite{Kasper29052012}. Figure~\ref{fig:comp1} shows the accuracy statistics of a reconstructed object. The average error as well as the standard deviation indicate that the quality of the models lies within the noise range of the sensor at the modelling distance. Moreover, the error distributions are comparable to those reported by~\cite{kehl2014bmvc} that uses a similar evaluation metric. Table~\ref{tab:results} shows further accuracy statistics.
\begin{table}[ht]
 \centering
   \begin{tabular}{|l|c|c|} \hline 
   & mean distance [mm] & $\sigma$ [mm]\\
   \hline\hline
   spray can & 2.16 & 1.53 \\ \hline
   box & 1.82 & 1.44 \\ \hline
   toy car & 1.71 & 1.96 \\ \hline
   \end{tabular}
   \caption{Comparison with laser scanner models.}
 \label{tab:results}
\end{table}

\section{Conclusion}

In this paper we have presented a flexible object reconstruction pipeline. Unlike most of the reconstruction and modelling tools out there, our proposal is able to reconstruct full 3D models of objects by changing the object configuration across different sessions. We have shown how the registration of different sessions can be carried out on featureless objects by exploiting the modelling setup where objects lie on a stable surface.
Another key functionality of our proposal is the ability to export object models in such a way that they can directly be used for object recognition and tracking. 
We believe that these tools will facilitate research in areas requiring object perception (e.g. human-object or robot-object interaction, grasping, object search as well as planning systems).

\section*{Acknowledgments}
The research leading to these results has received funding from the European Community Seventh Framework Programme FP7/2007-2013 under grant agreement No. 600623, STRANDS, No. 288146, HOBBIT and No. 610532, SQUIRREL and by Sparkling Science -- a programme of the Federal Ministry of Science and Research of Austria (SPA 04/84, FRANC).

\bibliography{root,hannes}

\end{document}